\documentclass{article}

\usepackage{ijcai23}

\usepackage[utf8]{inputenc}
\usepackage{makecell}
\usepackage{amsfonts}
\usepackage{amsmath}
\usepackage{booktabs}
\usepackage{graphicx}
\usepackage{natbib}

\title{Deep Anomaly Detection on Tennessee Eastman Process Data}
\author{Fabian Hartung$^{1,2}$ \and Billy Joe Franks$^1$ \and Tobias Michels$^1$ \and Dennis Wagner$^1$ \and Philipp Liznerski$^1$ \and Steffen Reithermann$^1$ \and Sophie Fellenz$^1$ \and Fabian Jirasek$^1$ \and Maja Rudolph$^3$ \and Daniel Neider$^4$ \and Heike Leitte$^1$ \and Chen Song$^5$ \and Benjamin Kloepper$^5$ \and Stephan Mandt$^6$ \and Michael Bortz$^7$ \and Jakob Burger$^8$ \and Hans Hasse$^1$ \And Marius Kloft$^{1,*}$
\affiliations
$^{1}$Technische Universität Kaiserslautern, Germany\\
$^{2}$BASF SE, Gas Treatment Technology, Germany\\
$^{3}$Bosch AI, USA\\
$^{4}$Technische Universität Dortmund, Germany\\
$^{5}$ABB Corporate Research Center Ladenburg, Germany\\
$^{6}$University of California Irvine, USA\\
$^{7}$Fraunhofer ITWM, Germany\\
$^{8}$Technische Universität München, Germany\\
Email corresponding author: kloft@cs.uni-kl.de\\
}
\date{\today}

\begin{document}

\maketitle

\begin{abstract}
This paper provides the first comprehensive evaluation and analysis of modern (deep-learning) unsupervised anomaly detection methods for chemical process data. We focus on the Tennessee Eastman process dataset, which has been a standard litmus test to benchmark anomaly detection methods for nearly three decades. Our extensive study will facilitate choosing appropriate anomaly detection methods in industrial applications.
\end{abstract}

\textbf{Keywords:} Anomaly detection, Chemical Process Data, Benchmark, Tennessee Eastman process, Time series

\section{Introduction}

Anomaly detection, i.e., detecting data that deviates from normality, is a fundamental method in machine learning and artificial intelligence. It is significant in many application domains, from detecting fake reviews in online shopping portals and bots in social networks to tumor detection and industrial fault detection. Anomaly detection is especially significant in safety-critical applications. While an undetected fake review in an online shopping portal may be harmless, failing to recognize anomalies in a chemical plant or a self-driving car may put lives at stake.

In chemical plants, most data is recorded during regular or problem-free operation—the normal data. Anomalies, in contrast, occur very rarely, and they can appear to the process or control engineers to be nominal behavior. Here, computing methodology naturally comes into play. Machine learning enables searching massive datasets and accurately detects anomalies, even when they are rare \citep{garg2017enclass}. 
There is a large body of literature on detecting anomalies in chemical processes using machine learning \citep{chadha2019comparison, monroy2009anomaly, song2019narrative}. Over the past three decades, the Tennessee Eastman process (TEP) has arisen as a litmus test for learning anomaly detection on chemical process data. Virtually any newly proposed method is benchmarked by default on the TEP dataset, originally recorded by \citet{downs1993plant} using a model TEP simulator for data generation. For the following survey a modiefied version will be used \citep{TEPDATA2022}.

However, except for some \citep{chadha2021deep,spyridon2018generative, plakias2022novel, yang2019generative, neuburgervariational, chadha2019comparison}, all papers evaluate shallow unsupervised anomaly detection methods (not including neural networks) on the TEP dataset. But shallow machine learning is not adequate for complex, structured data, such as the time series occurring in chemical plants and the TEP. On such data, most of the many seminal advances in artificial intelligence during the last decade have been enabled by deep neural networks.

In 2018, \citet{ruff2018deep} provided one of the earliest general-purpose deep learning approaches to anomaly detection.  The paper triggered a wave of follow-up work, resulting in the new field of ‘deep anomaly detection’ \citep{ruff2021unifying}. Over the past four years, the detection error of unsupervised anomaly detection methods has been reduced drastically, from 35\% (best shallow method, 2017) to 1\% (best deep method, 2021) on CIFAR-10-AD, a standard anomaly detection benchmark dataset \citep{ruff2018deep, liznerski2022exposing}. Since then, deep anomaly detection approaches have been widely adopted in industrial practice.
Most recent breakthroughs in modern anomaly detection have been achieved on image data. However, the data in chemical plants—and particularly the TEP—are time series. Time series exhibit intriguing temporal interdependencies, well-suited for deep learning. Very recently, the first deep anomaly detection methods on time-series data were introduced, and their high potential tested on various benchmarks \citep{qiu2021neural}. To date, there exist some 30 methods based on neural networks for anomaly detection on time series. 

However, the research on the TEP has not caught up yet with these highly significant advances in unsupervised deep anomaly detection on time series. There exists no compelling up-to-date comparison of modern methods, most of which have been developed within the last two years. Thus it is unclear which methods should ideally be used on such data to achieve maximal detection performance. Using inferior detection methods may lead to unnecessary errors or even put lives at risk when using them for real operation in plants. 

With the present work, we intend to change this. This paper evaluates and compares all 27 unsupervised deep anomaly detection methods for time series existing to date, regarding their detection accuracy on the TEP data. The analysis represents the first—and by far the most comprehensive—comparison of modern unsupervised anomaly detection methods on chemical process data. 
Our analysis also yields insights into which anomaly detection methods might be most suitable for application to real chemical process data. Establishing deep anomaly detection in real chemical processes would open the route for new, yet unexplored, ways to control these processes—with a perspective to advance autonomously running chemical processes. 

\section{Related Work}

Early papers on deep anomaly detection (AD) on times series were based on either reconstruction or forecasting objectives. Reconstruction approaches train an autoencoder (AE) on mostly normal training data so that the AE learns to compress and reconstruct normal data well. Samples not reconstructed well are considered anomalous. The deviation from the reconstruction to the input is the anomaly score \citep{hasan2016learning, luo2017remembering, malhotra2016lstm, mirza2018computer, zhang2019deep, audibert2020usad, thill2020time, kim2022towards, Hua2023, zhan2022stgat}. Forecasting models extrapolate a series's current and past data to predict future time steps. The anomaly score is the difference between the predicted and the actual future data \citep{malhotra2015long, filonov2016multivariate, munir2018deepant, he2019temporal, deng2021graph}. Typically, both reconstruction and forecasting methods reconstruct each time step and aggregate their anomaly scores for an anomaly score of the entire time series.

Another branch of AD methods is based on generative models such as variational autoencoders (VAEs) \citep{solch2016variational, xu2018unsupervised, park2018multimodal, guo2018multidimensional, su2019robust, li2020anomaly} and generative adversarial neural networks (GANs) \citep{zhou2019beatgan, li2019mad, niu2020lstm, geiger2020tadgan, sabokrou2018adversarially, liu2018future}. GANs jointly train two networks: a discriminator network to distinguish between accurate and generated data and a generator network to create samples that fool the discriminator. Anomaly scores are either based on the discriminator or are the deviation between the test sample and the best-fitting generated data sample. Some methods combine the above mentioned methods to get the best parts from all worlds \citep{said2020network, zhao2020multivariate}.

Inspired by the success of supervised classifiers, there is also a paradigm called "one-class classification" \citep{ruff2018deep}. This work trains a network to map normal samples to a hypersphere \citep{ruff2018deep} or hyperplane \citep{scholkopf2001estimating} and anomalous data away from them. This paradigm has recently been used for AD on time series \citep{said2020network, shen2020timeseries}. A more direct application of classifiers for AD requires anomalous training samples. Since AD is typically unsupervised, these samples are not available. One approach to solve this issue is using random internet data as auxiliary anomalies during training. This approach is termed outlier exposure and is successful on images \citep{liznerski2022exposing, hendrycks2018deep}. However, pertinent data is unavailable for time series, so \citet{goyal2020drocc} proposed to train a network to distinguish between normal training data and synthetically generated anomalies. The classifier's certainty for the anomalous class defines the anomaly score for test samples. The most recent approach to time series AD uses self-supervised learning \citep{qiu2021neural}. This method designs an auxiliary training objective. Normal data samples are transformed, and the network has to predict which type of transformation has been applied. Since, for anomalous data, a correct prediction will be difficult, the value of the method's decision certainty is the anomaly score for test samples.

\section{Benchmarking Deep Time-Series Anomaly Detection on TEP}

In this section, a more detailed explanation of the evaluation follows. First, we present the TEP data and explain the metrics used for the review. Finally, the implementation and evaluation protocol is presented.

\subsection{TEP dataset}

TEP was based on an existing plant and the processes running in it. The data itself is synthetic, i.e., a simulation of the plant. It consists of five main modules, each a two-stage reactor, a condenser, a vapor-liquid separator, a stripper, and a reboiler, as well as 11 pneumatic valves, two pumps, and a compressor \citep{manca2020tennessee}.

The version of the TEP data used here is available online \citep{TEPDATA2022} and is referenced in \citet{rieth2018issues}. In addition to error-free data on which the algorithms are to be trained, it contains 20 different types of erroneous data sets and their complete simulation. Of these 21 data sets, there are 500 other runs, each of which is initialized with a different random value. The time points in each sample are generated every three minutes for 25 hours for the training data and 48 hours for the test data with 53 parameters.

\subsection{Metrics}
To compare and evaluate the examined algorithms with each other, a metric is necessary that measures the quality of the methods. Work on AD uses different evaluation metrics depending on the data. Some metrics, like the F1-score, require a binary decision; i.e., model outputs in \{0, 1\} where 0 denotes normal and 1 anomalous. Others, like the receiver operator characteristic or precision-recall curve, work with continuous anomaly scores. For AD on time series, the F1-score and are under the precision-recall curve are the most commonly used metrics, which is why we evaluate the methods in this paper using both.  

An anomaly detector generates an anomaly score for each point in time of a time series. If this value exceeds a certain threshold, the respective method determines this point in time as an anomaly. The F1-score considers four options of evaluation for each time point: true positive (TP - a correctly detected anomaly), false negative (FN - an anomaly that was not detected), true negative (TN - a correctly identified normal point), and false positive (FP - a normal point mistakenly detected as an anomaly). With these four classes, two metrics can be calculated. One is precision, the proportion of TP among all detected anomalies (TP+FP), and the other is recall, the balance of TP anomalies among all true anomalies (TP+FN). Intuitively, precision describes the accuracy with which a detected anomaly is anomalous, and recall describes the accuracy with which the model detects true anomalies. The F1-score combines precision and recall in one metric, which can be calculated at every point of the time series:

\begin{align}
F1-Score &= 2 \frac{Precision * Recall}{Precision + Recall} \\
&= \frac{2 * TP}{2*TP+FN+FP}
\end{align}

These F1-scores are averaged over the whole time series to receive the total F1-Score.

The area under the precision-recall curve (AUPRC) can be used as a second metric for comparing methods. For every threshold, its respective recall and precision are calculated. As the threshold decreases, the recall increases to 1, which is plotted on the x-axis. The precision is plotted on the y-axis and can be arbitrary but generally decreases as the recall increases. The AUPRC measures the model's overall performance for any threshold. In essence, the higher the AUPRC, the higher the precision for any recall. In practice, there is a real-world cost associated with both FN and FP. Generally, the cost for undetected anomalies (FN) is higher than the cost of falsely detecting an anomaly (FP). However, the specific costs need to be defined case-by-case; therefore, the optimal threshold depends on the particular use case. The AUPRC is a good metric in case the specific costs are unknown since the higher AUPRC is, the lower these associated costs are expected to be.

\subsection{Evaluation and implementation}

For an equal and fair evaluation of the considered methods, all methods were implemented in the same Python environment and were trained and evaluated using PyTorch \citep{paszke2019pytorch}. Since some methods require an unlabeled validation set to adjust the parameters of the anomaly detector, a quarter of the training dataset was separated for this purpose. The test dataset was divided into five folds of equal size to adjust the hyperparameters of each method by optimizing them on each fold and evaluating the performance of the best model with the remaining folds. To avoid time dependencies, directly neighboring folds were excluded. Finally, all folds were averaged, the methods were compared using the best F1-score, and AUPRC received the best grid parameters. For better comparability, the size of the parameter grid of each method was chosen so that each one had a training and evaluation time of 24 hours. In total, the evaluation contains 27 methods listed below. As proposed from \citet{kim2022towards}, we added an Untrained-LSTM-AE as a baseline.

\subsection{Results}

Table 1 shows the experiments' results, implemented methods, and a reference to their original publications. The methods are ranked according to performance, and the results are rounded to four decimal places. The rankings are computed with the exact results. With few exceptions, both metrics and their associated rankings show similar results. It can only be observed for GMM-GRU-VAE, LSTM-AE-OC-SVM, and TCN-S2S-P differences of more than ten places in their order. The BeatGAN, TCN-S2S-AE, and Dense-AE methods score best. The weakest performers are GDN, LSTM-2S2-P, and THOC. It should be noted that Untrained-LSTM-AE, proposed above as a baseline, ends up in the upper midfield.

\begin{table*}[]
    \centering
    \resizebox{\linewidth}{!}{
    \begin{tabular}[bth]{lcccccc}
        \toprule
        \textbf{Method} & \textbf{Method Type} & \makecell{\textbf{F1-} \\ \textbf{Score}} & \makecell{\textbf{F1-} \\ \textbf{Score} \\ \textbf{Ranking}} & \textbf{AUPRC} & \makecell{\textbf{AUPRC} \\ \textbf{Ranking}} & \makecell{\textbf{Total} \\ \textbf{Ranking}} \\
        \midrule
        \textbf{BeatGAN} \citep{zhou2019beatgan} & Generative-GAN & 0.9699 & 1 & 0.9896 & 2 & 1 \\
        \midrule
        \textbf{TCN-S2S-AE} \citep{thill2020time} & Reconstruction & 0.9632 & 3 & 0.9914 & 1 & 2 \\
        \midrule
        \textbf{Dense-AE} \citep{audibert2020usad} & Reconstruction & 0.9631 & 4 & 0.9880 & 3 & 3 \\ 
        \midrule
        \textbf{LSTM-AE} \citep{malhotra2016lstm} & Reconstruction & 0.9506 & 5 & 0.9861 & 4 & 4 \\
        \midrule
        \textbf{LSTM-P} \citep{malhotra2015long} & Forecasting & 0.9693 & 2 & 0.9824 & 8 & 5 \\
        \midrule
        \textbf{MSCRED} \citep{zhang2019deep} & Reconstruction & 0.9353 & 7 & 0.9842 & 5 & 6 \\
        \midrule
        \textbf{Donut} \citep{xu2018unsupervised} & Generative-VAE & 0.9450 & 6 & 0.9829 & 7 & 7 \\
        \midrule
        \textbf{LSTM-VAE} \citep{solch2016variational} & Generative-VAE & 0.9334 & 11 & 0.9831 & 6 & 8 \\
        \midrule
        \textbf{OmniAnomaly} \citep{su2019robust} & Generative-VAE & 0.9336 & 9 & 0.9808 & 12 & 9 \\
        \midrule
        \textbf{SIS-VAE} \citep{li2020anomaly} & Generative-VAE & 0.9335 & 10 & 0.9790 & 14 & 10 \\
        \midrule
        \textbf{Untrained-LSTM-AE} \citep{kim2022towards} & Reconstruction & 0.9333 & 13 & 0.9792 & 13 & 11 \\
        \midrule
        \textbf{LSTM-DVAE} \citep{park2018multimodal} & Generative-VAE & 0.9333 & 16 & 0.9811 & 11 & 12 \\
        \midrule
        \textbf{USAD} \citep{audibert2020usad} & Reconstruction & 0.9333 & 12 & 0.9779 & 16 & 13 \\
        \midrule
        \textbf{GMM-GRU-VAE} \citep{guo2018multidimensional} & Generative-VAE & 0.9291 & 21 & 0.9815 & 10 & 14 \\
        \midrule
        \textbf{TCN-S2S-P} \citep{he2019temporal} & Forecasting & 0.9172 & 23 & 0.9821 & 9 & 15 \\
        \midrule
        \textbf{LSTM-MAX-AE} \citep{mirza2018computer} & Reconstruction & 0.9333 & 18 & 0.9786 & 15 & 16 \\
        \midrule
        \textbf{LSTM-AE-OC-SVM} \citep{said2020network} & Hybrid & 0.9337 & 8 & 0.9511 & 26 & 17 \\
        \midrule
        \textbf{LSTM-VAE-GAN} \citep{niu2020lstm} & Generative-GAN & 0.9333 & 14 & 0.9735 & 20 & 17 \\
        \midrule
        \textbf{GenAD} \citep{Hua2023} & Reconstruction & 0.9333 & 19 & 0.9755 & 19 & 19 \\
        \midrule
        \textbf{TadGAN} \citep{geiger2020tadgan} & Generative-GAN & 0.9333 & 15 & 0.9690 & 23 & 19 \\
        \midrule
        \textbf{STGAT-MAD} \citep{zhan2022stgat} & Reconstruction & 0.9267 & 22 & 0.9767 & 17 & 21 \\
        \midrule
        \textbf{Mad-GAN} \citep{li2019mad} & Generative-GAN & 0.9333 & 17 & 0.9621 & 24 & 22 \\
        \midrule
        \textbf{MTAD-GAT} \citep{zhao2020multivariate} & Hybrid & 0.9097 & 25 & 0.9758 & 18 & 23 \\
        \midrule
        \textbf{DeepANT/TCN-P} \citep{munir2018deepant} & Forecasting & 0.9114 & 24 & 0.9712 & 22 & 24 \\
        \midrule
        \textbf{GDN} \citep{deng2021graph} & Forecasting & 0.9078 & 26 & 0.9722 & 21 & 25 \\
        \midrule
        \textbf{LSTM-2S2-P} \citep{filonov2016multivariate} & Forecasting & 0.9327 & 20 & 0.9171 & 27 & 25 \\
        \midrule
        \textbf{THOC} \citep{shen2020timeseries} & Hybrid & 0.9074 & 27 & 0.9618 & 25 & 27 \\
        \bottomrule
    \end{tabular}}
    \caption{This table shows the performance of all evaluated methods. For each method, the table lists its reference, the achieved best F1-score, and best AUPRC. The table also lists the ranking according to F1-score, AUPRC, and their mean. The methods are sorted according to the best mean of F1-score and AUPRC.}
    \label{tab:my_label}
\end{table*}

\section{Discussion and Conclusion}

Even though a generative model was ranked first in these experiments, you can conclude that the reconstruction methods performed best on average, followed by the forecasting and, finally, the generative models. Even the proposed baseline, which belongs to the reconstruction methods, achieved an above-average ranking.

For future work, a few more things need to be investigated. On the one hand, it has to be considered that the TEP data are synthetic. Despite the simulation's quality, chemical processes are multifaceted, and, especially with real data, other parameters may play a role that cannot be simulated this way. All methods have yielded high scores. That could be related to the studied synthetic data with defined synthetic faults introduced in a fault-free run. The task will be considerably more challenging for actual chemical process data, but the present study is a starting point to tackle this problem. The challenge here will be in uncovering the data and correctly labeling the anomalies in that data. On the other hand, additional metrics should be taken into account. The F1-score and AUPRC are a reasonable basis for comparison but cannot assess longer periods and interdependent points, as with time series \citep{kim2022towards, doshi2022tisat}.

The benchmarking in this paper can guide further research and practitioners in selecting a suitable method for anomaly detection on chemical time series.

\clearpage
\appendix


\section*{Acknowledgement}
Part of this work was conducted within the DFG research unit FOR 5359 on Deep Learning on Sparse Chemical Process Data and the BMWK project KEEN (01MK20014U,01MK20014L).

\bibliographystyle{named}
\bibliography{refs}

\clearpage
\section{Supporting Information: Methods}
To identify a data point as normal or anomaly, the core element of a method is calculating an anomaly score. In most cases, this involves evaluating the current point using only knowledge about the previous points and, in some cases, using individual time windows. The following is an overview of the most relevant methods for calculating the anomaly score in recent years and shows the implementations of the methods in this evaluation.

\subsection{Reconstruction-based methods}
The basic idea of autoencoders (AE) is to project input data onto a latent space of lower dimension and then project it back into the input space. In this process, the network must learn to preserve the information in the best possible way but cannot learn an identity function due to the dimensionality differences of the spaces. Since the training is done exclusively on normal data, the projection of an anomaly should produce a more significant reproduction error. They can use the mean squared error as the anomaly score and the training itself.

\textbf{LSTM-AE}	\citep{malhotra2016lstm} uses an LSTM network as an encoder and decoder. The encoder LSTM gives its final hidden state to the decoding LSTM as an initial hidden state and reconstructs the input in reverse order. It uses real data as inputs in training and its predictions during testing.

\textbf{LSTM-Max-AE} \citep{mirza2018computer} proposes some changes by using the mean or maximum of the hidden states of the encoder. During reconstruction, the latent representation is used as an input for all time steps. Here, the inputs are reconstructed in the same order.

\textbf{MSCRED} \citep{zhang2019deep} doesn't use the raw inputs but creates signature matrices by capturing the correlation of time series segments. A fully 2D-convolutional network is applied, and the output is fed into a 2D-convolutional LSTM encoder and decoder.

\textbf{USAD} \citep{audibert2020usad} introduced two AE with a shared encoder. The training splits into two phases: First, train both AE to minimize the reconstruction error. Second, let the AE compete against each other. At the same time, the second AE aims to distinguish actual samples from those generated by the first AE, which tries to fool the other. A combination of reconstruction and adversarial loss yields the anomaly score for each point.

\textbf{Dense-AE} \citep{audibert2020usad} is the same fully-connected encoder and decoder from USAD but used in a regular AE without the second decoder and additional adversarial objective. The MSE between input and reconstruction is used for both the anomaly score and the training loss.

\textbf{TCN-S2S-AE} \citep{thill2020time} proposes a temporal convolutional network (TCN) in the encoder and a transposed TCN in the decoder. This fully convolutional AE architecture uses the LogCosh loss as a training objective. They propose to fit a Gaussian on the errors over the test set during testing. So, this method is unusable in online settings. To be comparable to other methods in this context, the Gaussian is fitted to a held-out validation set.

\textbf{Untrained-LSTM-AE} \citep{kim2022towards} is proposed as a baseline for better comparison to newly developed methods. Therefore, an untrained autoencoder with a single-layer LSTM is used for that. The initialization is random.

\textbf{GenAD} \citep{Hua2023} proposes to split an input time series into five folds of equal size. Then it selects 20\% of N-dimensions to be masked in one fixed fold. The left 80\% unmasked dimensions in this fold and all dimensions in the other four folds are then used to reconstruct the 20\% masked series. Since this choice is random, the model needs to learn correlations and temporal patterns to minimize the loss. The implementation masks each feature in the input time series once and lets the model compute its reconstruction. The reconstruction error is measured by the LogCosh metric and considers the feature anomalous when exceeding some threshold. The entire time series is considered anomalous at a certain point if more than a predetermined fraction of the input features is anomalous at this point.

\textbf{STGAT-MAD} \citep{zhan2022stgat} applies several 1D-convolutional layers with varying kernel sizes on an input time series. The resulting sequences are passed parallelly through several graph attention and convolutional layers. Afterward, their concatenated outputs are fed to a bi-LSTM decoder attempting to reconstruct the input. The squared error is used for both training loss and anomaly score. 

\subsection{Forecasting-based methods}
Instead of reconstructing a given input and measuring its quality, another method to detect anomalies is to predict the next time series step(s). These predictions can be compared with the following original time series steps. The point is marked anomalous if the difference exceeds a specified threshold, calculated mainly by the MSE or the mean absolute error (MAE). The number of predicted steps $k \geq 1$ is called the prediction horizon. By training these methods on normal data, the networks should be able to give well predictions to normal test data and produce higher prediction errors for anomalous data.

\textbf{LSTM-P} \citep{malhotra2015long} proposes using a multilayer LSTM to extract features and generate l-steps predictions with an FC NN. They use MSE loss for training and fit a multivariate Gaussian to the errors of the held-out validation set. After learning the distribution, the anomaly score corresponds to the negative log-likelihood.
LSTM-2S2-P [28] uses a multilayer LSTM similarly but predicts the forecast with the hidden features at each time step. By doing this, the model is a sequence-to-sequence predictor, and the anomaly score is yielded by an exponentially weighted moving average of the reconstruction error. 

\textbf{DeepANT/TCN-P} \citep{munir2018deepant} chains a max pooling TCN with an MLP in a row to predict the following k points from the input window w. The model is trained with MAE and the anomaly score yield by MSE between a prediction and its original time series point. If $k \geq 1$, the average of all predictions for a single time step is used to calculate the anomaly score with MSE.

\textbf{TCN-S2S-P} \citep{he2019temporal} proposes to pass the input window through a dilated causal TCN. The outputs of the last three layers along the feature dimension are concatenated and given to a final convolutional layer, kernel size one, and D filters. By doing this, the output is a size w x D window and is shifted by one step. Again, the MSE loss is used during training, and a Gaussian distribution is fitted to the prediction errors. Only the last point in the prediction window can be used for this method in an online setting.

\textbf{GDN} \citep{deng2021graph} builds a graph with features as nodes and edges as relations between features. An Embedding vector for each feature is trained and directed edges from each feature to the top $m \in \mathbb{N}$ features based on cosine similarity between the feature embeddings. The graph is dynamically recreated for each input batch. The prediction is yielded by applying a graph attention mechanism (Petar Vel\u{\i}ckovi'c, Graph attention networks 2018) and passing the outputs- to an MLP. The authors' way of calculating the anomaly score as two statistics over the test set, MSE for training, and MAE for anomaly score, make GDN an offline method. The unscaled MSE is used as the anomaly score to change that to an online use case.

\subsection{Generative Methods}
Generative methods model the data-generating distribution directly. They train a generative model on some latent space with a predefined prior, producing samples close to the real data. Usually, those models offer some way of computing the marginal likelihood of a data point under the model they learned, which can be used to derive anomaly scores.

\subsubsection{VAE-Based Methods}

\textbf{LSTM-VAE} \citep{solch2016variational} sets likelihood and the posterior approximation to be Gaussian and chooses all NNs to be single-layer LSTMs. In each time step, the encoder returns a mean and covariance component. They produce $\mu = (\mu_1, …, \mu_t)$ by another LSTM and use a Gaussian normal distribution with that mean and the identity matrix as covariance matrix as a prior. The anomaly score is calculated with the negative ELBO.

\textbf{Donut} \citep{xu2018unsupervised} uses MLPs as encoders and decoders. They set some time steps in the input to zero to mask them and train by maximizing a modified version of ELBO that accounts for the input masking. As an anomaly score, they propose the so-called "reconstruction probability," although combining it with elaborate mechanisms to reconstruct missing data. Since the TEP data do not include missing data, this is irrelevant to this work. Since the original version of Donut only supports univariate time series, extend it to a multivariate case by applying MLPs to the flattened multivariate input window and masking only random features in random time steps instead of entire-time steps. 

\textbf{LSTM-DVAE} \citep{park2018multimodal} does the same as LSTM-VAE with three changes. First, apply zero-mean Gaussian noise to any input. Second, computing their prior mean for each time step as
\begin{equation}
\mu_t=\left(1-\frac{t}{T}\right)v_1+\frac{t}{T}v_T
\end{equation}
	With learnable parameters $v_1, v_T \in \mathbb{R^D}$. Finally, as an anomaly score, they use the reconstruction probability.
 
\textbf{GMM-GRU-VAE} \citep{guo2018multidimensional} chooses GRUs for their encoder and decoder. For their variational posterior approximation, they use a Gaussian mixture distribution with K components and a Gaussian mixture with learnable parameters for each component. The chosen anomaly score is the reconstruction probability.  

\textbf{OmniAnomaly} \citep{su2019robust} also uses an encoder and decoder on a GRU basis. The encoder defines parameters for multivariate normal distribution. After sampling latent variable z from it, they apply a planar normalizing flow. For the prior is a Kalman filter, a linear Gaussian state space model, chosen. The reconstruction probability is again used as the anomaly score.

\textbf{SIS-VAE} \citep{li2020anomaly} proposes a GRU-based VAE to reconstruct smooth time series. They add a KL-divergence term to the ELBO between adjacent time steps. This encourages the distribution of the predicted time series for two close points to be similar. Usually, the reconstruction probability is used as an anomaly score. 

\subsubsection{GAN-based methods}

\textbf{BeatGAN} \citep{zhou2019beatgan} uses a TCN-based AE as the generator, training a minimization of the MSE between input and its reconstruction and the MSE between their feature maps in the discriminator's second-to-last layer. A TCN-based discriminator is trained on the standard GAN loss. As an anomaly score, the MSE of the AE is used.

\textbf{Mad-GAN} \citep{li2019mad} uses a GAN-based approach LSTMs as both a generator and discriminator. In addition to the usual discriminator score, they also use a reconstruction score. Starting with a latent variable and passing it through the generator, they use a Gaussian/RBF kernel to compute the similarity between this generated sample and the current original input. They use the difference between 1 and this similarity as a reconstruction error. With gradient-based methods, they minimize this error down to a certain threshold. The MAE yielded the anomaly score between the reconstructed and original input and the discriminator's output.

\textbf{LSTM-VAE-GAN} \citep{niu2020lstm} set up a GAN using a decoder of an LSTM-based VAE as the generator and an LSTM as the discriminator. The original and reconstructed sequences are passed through all but the last layer of the discriminator. Since the discriminator should also be able to detect transformed samples generated from the posterior approximation, besides the ones from the standard normal distribution, its loss has an additional term to detect these samples. The anomaly score is, together with the negative discriminator score, a convex combination of the MAE between x and its reconstruction. 

\textbf{TadGAN} \citep{geiger2020tadgan} uses bidirectional LSTMs for an AE. Considering these decoders and encoders as generators for two Wasserstein GANs, one GAN uses the decoder LSTM as its generator. This maps random samples from an ordinary standard distribution to the input data space. A TCN-based discriminator learns to decide whether the data was a real input or a generated sample. The encoder LSTM is the generator of the second GAN mapping data points to the latent space. This GAN's TCN discriminator must now decide if its input is an encoded data point or a random sample from the standard normal distribution. The loss function contains the reconstruction error of the AE measured by MSE.
Additionally, the discriminator score is calculated and normalized by their means and standard deviations in the test set. The final anomaly score is a convex combination of both absolute values. This method is turned into an online method by computing the statistics of both scores on a held-out part of the training set instead.

\subsection{Hybrid Methods}
Some methods share principles of different classes from above and combine them in new ways. This is a list of these Hybrid methods. 

\textbf{LSTM-AE OC-SVM} \citep{said2020network} proposes an AE with multilayer LSTMs. Typically, the anomaly score is based on the reconstruction error of an AE, but the authors train an OC-SVM \citep{scholkopf2001estimating} on the latent vectors. These are produced by applying the encoder to the held-out clean validation set instead. To make this method more comparable, return the raw scores, i.e., signed distances instead of predictions, and the same architecture as for LSTM-Max-AE.

\textbf{MTAD-GAT} \citep{zhao2020multivariate} use two graph attention modules (Petar Veliˇckovi'c, Graph attention networks 2018) and applies them on top of a TCN. One takes features as nodes, and the other time points in a window as nodes. Their output is concatenated and fed into a GRU. They use a fully connected graph as the input. The final hidden state serves as the new input for an MLP for the prediction of the next time point and, simultaneously, as the latent variable for a VAE with an MLP decoder. The model is trained by additively combining the MSE of the prediction and the VAE's ELBO loss. MSE and the reconstruction probability of the VAE under the usage of a trade-off coefficient between 0 and 1 are combined.

\end{document}